\documentclass[times,review,nopreprintline]{elsarticle}

\usepackage{framed,multirow}

\usepackage{amssymb}
\usepackage{latexsym}

\usepackage{url}
\usepackage{xcolor}

\usepackage{hyperref}

\usepackage{acro}
\usepackage{comment}
\usepackage{appendix}
\usepackage{booktabs}
\usepackage{array}
\usepackage{caption} 
\usepackage{amsmath}
\usepackage{algorithm}
\usepackage[noend]{algpseudocode}

\newcommand{\new}[1]{\textcolor{black}{#1}}
\newcommand{\newtwo}[1]{\textcolor{black}{#1}}

\DeclareAcronym{CXR}{
 short = CXR ,
 long = chest X-ray
}

\DeclareAcronym{PACS}{
short=PACS,
long=picture archiving and communication system
}
\DeclareAcronym{PLCO}{
 short = PLCO,
 long = {Prostate, Lung, Colorectal and Ovarian} 
}
\DeclareAcronym{AUC}{
 short = AUC,
 long = area under the curve 
}
\DeclareAcronym{CAD}{
 short = CAD,
 long = computer-aided diagnosis 
}
\DeclareAcronym{HMLC}{
 short = HMLC,
 long = hierarchical multi-label classification 
}

\DeclareAcronym{PU}{
 short = PU,
 long = positive and unlabelled
}

\DeclareAcronym{HLCP}{
 short = HLCP,
 long = hierarchical label conditional probability
}

\DeclareAcronym{HLUP}{
 short = HLUP,
 long = hierarchical label unconditional probability
}

\DeclareAcronym{NCI}{
 short = NCI,
 long = National Cancer Institute
}

\DeclareAcronym{BR}{
 short = BR,
 long = binary relevance 
}
\DeclareAcronym{CE}{
 short = CE,
 long = cross entropy 
}
\DeclareAcronym{AP}{
 short = AP,
 long = average precision
}

\DeclareAcronym{CNN}{
 short = CNN,
 long = convolutional neural network
}

\DeclareAcronym{RNN}{
 short = RNN,
 long = recurrent neural network
}

\DeclareAcronym{NLP}{
short=NLP,
long=natural language processing
}

\newcolumntype{C}[1]{>{\centering\let\newline\\\arraybackslash\hspace{0pt}}m{#1}}
\newcommand{\eg}{\emph{e.g.},}
\newcommand{\ie}{\emph{i.e.},}

\newcommand{\Sec}{Section}
\newcommand{\Fig}{Figure}
\newcommand{\Tab}{Table}

\definecolor{newcolor}{rgb}{.8,.349,.1}

\journal{Medical Image Analysis}

\begin{document}


\begin{frontmatter}

\title{Deep Hiearchical Multi-Label Classification Applied to Chest X-Ray Abnormality Taxonomies}%



\author[1]{Haomin Chen \corref{cor1}}
\ead{hchen135@jhu.edu}
\author[2]{Shun Miao }
\author[3]{Daguang Xu }
\author[1]{Gregory D. Hager}
\author[2]{Adam P. Harrison \corref{cor1}}
\cortext[cor1]{Corresponding authors. Work performed at NVIDIA}
\ead{adam.p.harrison@gmail.com}
\address[1]{Johns Hopkins University, Baltimore, MD, US}
\address[2]{PAII Inc., Bethesda, MD, US}
\address[3]{NVIDIA AI-Infra, Bethesda, MD, US}


\begin{abstract}
\Acp{CXR} are a crucial and extraordinarily common diagnostic tool, leading to heavy research for \ac{CAD} solutions. However, both high classification accuracy \emph{and} meaningful model predictions that respect and incorporate clinical taxonomies are crucial for \ac{CAD} usability. To this end, we present a deep \ac{HMLC} approach for \ac{CXR} \ac{CAD}. Different than other hierarchical systems, we show that first training the network to model conditional probability directly and then refining it with unconditional probabilities is key in boosting performance. In addition, we also formulate a numerically stable cross-entropy loss function for unconditional probabilities that provides concrete performance improvements. Finally, we demonstrate that \ac{HMLC} can be an effective means to manage missing or incomplete labels. To the best of our knowledge, we are the first to apply \ac{HMLC} to medical imaging \ac{CAD}. We extensively evaluate our approach on detecting abnormality labels from \newtwo{the \ac{CXR} arm} of the \ac{PLCO} dataset, which comprises over $198,000$ manually annotated \acp{CXR}. When using complete labels, we report a mean \ac{AUC} of $0.887$, the highest yet reported for this dataset. These results are supported by ancillary experiments on the PadChest dataset, where we also report significant improvements, $1.2\%$ and $4.1\%$ in \ac{AUC} and \acl{AP}, respectively over strong ``flat'' classifiers. Finally, we demonstrate that our \ac{HMLC} approach can much better handle incompletely labelled data. These performance improvements, combined with the inherent usefulness of taxonomic predictions, indicate that our approach represents a useful step forward for \ac{CXR} \ac{CAD}.
\end{abstract}


\end{frontmatter}



\acresetall
\section{Introduction}
\Acp{CXR} account for a large proportion of ordered image studies, \eg{} in the US it accounted for almost half of ordered studies in 2006~\citep{Mettler_2009}. Commensurate with this importance, \ac{CXR} \emph{\ac{CAD}} has received considerable research attention, both prior to the popularity of deep learning~\citep{Jaeger_2013}, and afterwards~\citep{Wang_2017,yao2017learning,guendel_2019,Irvin_2019,Aurelia_Bustos_2019}. \newtwo{These efforts have met success and typically approach the problem as a standard multi-label classification scenario, which attempts to make a set of individual binary predictions for each disease pattern under consideration.} Yet, pushing raw performance further will likely require models that depart from standard multi-label classifiers. For instance, despite their importance to clinical understanding and interpretation~\citep{stevens_using_2007,humphreys_umls_1993,stearns_snomed_2001}, taxonomies of disease patterns are not typically incorporated into \ac{CXR} \ac{CAD} systems, or for other medical \ac{CAD} domains for that matter. This observation motivates our work, which uses \emph{\ac{HMLC}} to both push raw \ac{AUC} performance further and also to provide more meaningful predictions that leverage clinical taxonomies. 

Organizing diagnoses or observations into ontologies and/or taxonomies is crucial within radiology, \eg{} RadLex~\citep{Langlotz_2006}, with \ac{CXR} interpretation being no exception~\citep{Folio_2012,Fushman_2015,Dimitrovski_2011}. This importance should also be reflected within \ac{CAD} systems. For instance, when uncertain about fine-level predictions, \eg{} \emph{nodules} vs. \emph{masses}, a \ac{CAD} system should still be able to provide meaningful parent-level predictions, \eg{} \emph{pulmonary nodules and masses}. This parent prediction may be all the clinician is interested in anyway. Another important benefit is that observations are conditioned upon their parent being true, allowing fine-level predictors to focus solely on discriminating between siblings rather than on having to discriminate across all possible conditions. This can help improve classification performance~\citep{Bi_2015}.

Elegantly addressing the problem of incompletely labelled data is another benefit of incorporating taxonomy. To see this, note that many \ac{CXR} datasets are collected using \ac{NLP} approaches applied to hospital \acp{PACS}~\citep{Wang_2017,Irvin_2019}. This is a trend that will surely increase given that \acp{PACS} remain the most viable source of large-scale medical data~\citep{Kohli_2017,harvey2019standardised}. In such cases, it may not always be possible to extract fine-grained labels with confidence. For instance, imaging conditions may have only allowed a radiologist to report ``opacity'', instead of a more specific observation of ``infiltration'' vs. ``atelectasis''. Added to this inherent uncertainty is the fact that \ac{NLP} approaches for \ac{CXR} label extraction themselves can suffer from considerable levels of error and uncertainty~\citep{Irvin_2019, Erdi_2019}. As a result, it is likely that \ac{CAD} systems will increasingly be faced with incompletely labelled data, where data instances may be missing fine-grained labels, but still retain labels higher up in the clinical taxonomy. An \ac{HMLC} approach can naturally handle such incompletely labelled data.

For these reasons, we present a deep \ac{HMLC} approach for \ac{CXR} \ac{CAD}. We extensively evaluate our \ac{HMLC} approach on \newtwo{the \ac{CXR} arm of} the \emph{\ac{PLCO}} dataset~\citep{Gohagan_2000} with supporting experiments on the PadChest dataset~\citep{Aurelia_Bustos_2019}. Experiments demonstrate that our \ac{HMLC} approach can push raw performance higher compared to both leading ``flat'' classification baselines and other \ac{HMLC} alternatives. We also demonstrate that our \ac{HMLC} approach can robustly handle extremely large proportions of incompletely labelled data with much less performance loss than alternatives. To the best of our knowledge, we are the first to outline an \ac{HMLC} \ac{CAD} system for medical imaging and the first to characterize performance when faced with incompletely labelled data.

\subsection{Related Work}

\noindent
\textbf{CXR Classification:}
Because more than one abnormality can be observed on a \ac{CXR} at the same time, a \ac{CAD} \ac{CXR} system must operate in a multi-label setting. \newtwo{This is in contrast to \emph{multi-class} approaches, which typically attempt to make a single $n$-ary prediction per image.} Truly large-scale \ac{CXR} classification started with the CXR14 dataset and the corresponding model~\citep{Wang_2017}, with many subsequents improvements both in modeling and in dataset collection~\citep{Irvin_2019, Aurelia_Bustos_2019,johnson_mimic-cxr_2019}. These improvements include incorporating ensembling~\citep{islam_abnormality_2017}, attention mechanisms~\citep{guan_diagnose_2018, wang_chestnet:_2018,liu_sdfn:_2019}, and localizations~\citep{yan_weakly_2018,li_thoracic_2018,liu_sdfn:_2019,guendel_2019,Cai_2018}. Similar to \citep{guendel_2019}, we also train and test on the \ac{PLCO} dataset. However, \citep{guendel_2019} boosted their performance by incorporating the CXR14 dataset~\citep{Wang_2017} and a multi-task framework that also predicted the rough locations and the lung and heart segmentations. While the contributions of these cues, when available, are important to characterize and incorporate, our \ac{HMLC} approach can achieve higher \acp{AUC}\footnote{With the caveat of using different data splits, since there is no official split.} without extra data or auxiliary cues.

A commonality between these prior approaches is that they typically treat each label as an independent prediction, which is commonly referred to as \ac{BR} learning within the multi-label classification field~\citep{zhang_review_2014}. However, prior work has well articulated the limitations of \ac{BR} learning~\citep{Dembczynski_2012}. A notable exception to this trend is \citep{yao2017learning}, which modeled correlations between labels using a recurrent neural network. In contrast, our \ac{HMLC} system takes a different approach by incorporating top-down knowledge to model the conditional dependence of children labels upon their parents. In this way, we make predictions conditionally independent rather than globally independent, allowing the model to focus on discriminating between siblings rather than across all disease patterns. 

\noindent
\textbf{Hierarchical Classification:}
Given its potential to improve performance, incorporating taxonomy through hierarchical classification has been well-studied. Prior to the emergence of deep learning, seminal approaches used hierarchical and multi-label generalizations of classic algorithms~\citep{mccallum_improving_1998,cesa-bianchi_incremental_nodate,cai_exploiting_nodate,vens_decision_2008}. With the advent of deep learning, a more recent focus has been on adapting deep networks, typically \acp{CNN}, for hierarchical classification~\citep{Redmon_2017,Roy_2018,Yan_2014,guo_cnn-rnn:_2018,kowsari_hdltex:_2017}. Interestingly, \citep{cesa-bianchi_incremental_nodate} use an approach similar to popular approaches seen in more recent deep hierarchical \textit{multi-class} classification of natural images~\citep{Redmon_2017,Roy_2018,Yan_2014}, \ie{} train classifiers to predict conditional probabilities at each node.  Our approach is similar to these more recent deep approaches, except that we focus on \emph{multi-label} classification and we also formulate a numerically stable unconditional probability fine-tuning step. 

Other deep approaches used complicated combinations of \acp{CNN} and \acp{RNN}~\citep{guo_cnn-rnn:_2018,kowsari_hdltex:_2017}, but for our \ac{CXR} application we show that a much simpler approach that uses a shared trunk network for each of the output nodes can, on its own, provide important performance improvements over ``flat`` classifiers.

Within medical imaging, there is work on \ac{HMLC} medical image retrieval using either nearest-neighbor or multi-layer perceptrons~\citep{Pourghassem_2008} or decision trees~\citep{Dimitrovski_2011}. However, hierarchical classifiers have not received much attention for medical imaging \emph{\ac{CAD}} and deep \ac{HMLC}  approaches have not been explored at all. Finally, we note that the process of producing a set of binary \ac{HMLC} labels, given a set of pseudo-probability predictions, is a surprisingly rich topic~\citep{Bi_2015}, but here we focus on producing said predictions.

\noindent
\textbf{Incompletely Labelled Data:}
As mentioned, another motivating factor for \ac{HMLC} is its ability to handle incompletely or partially labelled data. Within the computer vision and text mining literature, there is a rich body of work on handling partial labels
~\citep{yu_large-scale_nodate,kong_large-scale_2014,zhao_semi-supervised_nodate,elkan_learning_2008,liu_building_2003,qi_mining_2011,bucak_multi-label_2011,yang_multi-instance_nodate}. When missing labels are positive examples, this problem has also been called \ac{PU} learning. Seminal \ac{PU} works focus on multi-class learning~\citep{elkan_learning_2008,liu_building_2003}. There are also efforts for \emph{multi-label} \ac{PU} learning~\citep{kong_large-scale_2014,yu_large-scale_nodate,zhao_semi-supervised_nodate,qi_mining_2011,bucak_multi-label_2011,yang_multi-instance_nodate}, which attempt to exploit label dependencies and correlations to overcome missing annotations. However, many of these approaches do not scale well with large-scale data~\citep{kong_large-scale_2014}. 

\citep{yu_large-scale_nodate} and \citep{kong_large-scale_2014} provide two exceptions to this, tackling large-scale numbers of labels and data instances, respectively. In our case, we are only interested in the latter, as the number of observable \ac{CXR} disease patterns remains manageable. We are able to take advantage of a hierarchical clinical taxonomy to model label dependencies, allowing us to avoid complex approaches to learn these dependencies, such as the stacking methods used by \citep{kong_large-scale_2014}. In this way, our approach is similar to that of \citep{cesa-bianchi_incremental_nodate}, who also use a hierarchy to handle \ac{PU} data through an incremental linear classification scheme. However, our approach uses deep \acp{CNN} and we are the first to show how \ac{HMLC} can help address the \ac{PU} problem for \ac{CAD} and the first to characterize performance of \ac{CXR} classifiers under this scenario.

\subsection{Contributions}
 Based on the above, the contributions of our work can be summarized as follows:
 \begin{itemize}
     \item Like other deep hierarchical \emph{multi-class} classifiers, we train a classifier to predict conditional probabilities. However, we operate in the \emph{multi-label} space and we also demonstrate that a second fine-tuning stage, trained using unconditional probabilities, can boost performance for \ac{CXR} classification even further.
     \item  To handle the unstable multiplication of prediction outputs seen in unconditional probabilities we introduce and formulate a numerically stable and principled loss function.
     \item Using our two-stage approach, we  are  the  first  to  apply \acf{HMLC}  to  \ac{CXR} \ac{CAD}. Our straightforward, but effective, \ac{HMLC} approach results in the highest mean \ac{AUC} value yet reported for the \ac{PLCO} dataset. 
     \item In addition, we demonstrate how \ac{HMLC} can serve as an effective means to handle incompletely labelled data. We are the first to characterize \ac{CXR} classification performance under this scenario, and experiments demonstrate how \ac{HMLC} can garner even greater boosts in classification performance. 
 \end{itemize}
 Finally, we note that portions of this work were previously published as a conference proceeding~\citep{chen_deep_2019}. This work adds several contributions: (1) we significantly expand upon the literature review; (2) we include the derivation of the numerically stable unconditional probability loss within the main body and have made its derivation clearer; (3) we include additional results with the PadChest~\cite{Aurelia_Bustos_2019} dataset to further validate our approach; and (4) we add the motivation, discussion, and experiments on incompletely labelled data. 
 
 
\section{Materials and Methods}

We introduce a two-stage method for \ac{CXR} \ac{HMLC}. We first outline the datasets and taxonomy we use in Section~\ref{sec:datasets} and then overview the general concept of \ac{HMLC} in Section~\ref{sec:hmlc}. This is followed by Sections~\ref{sec:conditional} and \ref{sec:unconditional}, which detail our two training stages that use conditional probability and a numerically stable unconditional probability formulation, respectively.

\subsection{Datasets and Taxonomy}
\label{sec:datasets}
The first step in creating an \ac{HMLC} system is to create the label taxonomy. In this work, our main results focus on the labels and data found within the \ac{CXR} arm of the \ac{PLCO} dataset~\citep{Gohagan_2000}, a large-scale lung cancer screening trial  that collected $198\,000$ \acp{CXR} with image-based annotations of abnormalities obtained from multiple US clinical centers. While other large-scale datasets~\citep{Wang_2017,Aurelia_Bustos_2019, Irvin_2019,johnson_mimic-cxr_2019} are \emph{extraordinarily valuable}, their labels are generated by using \ac{NLP} to extract mentioned disease patterns from radiological reports found in hospital \acp{PACS}. \newtwo{While medical \ac{NLP} has made great strides in recent years, it still remains an active field of research, \eg{} NegBio still reports limitations with uncertainty detection, double-negation, and missed positive findings for certain \ac{CXR} terms~\citep{peng_2017}. However, irrespective of the \ac{NLP}'s level of accuracy, there are more inherent limitations to using text-mined labels. Namely, examining a text report is no substitute for visually examining the actual radiological scan, as the text of an individual report is not a complete description of the \ac{CXR} study in question. Thus, terms may not be mentioned, \eg{} ``no change'', even though they are indeed visually apparent. Additionally, a radiologist will consider lab tests, prior radiological studies, and the patient's records when writing up a report. Thus, mentioned terms, and their meaning, may well be influenced by factors that are not visually apparent. Compounding this, text which is unambiguous given the patient's records and radiological studies may be highly ambiguous when only considering text alone, \eg{} whether a pneumothorax is untreated or not~\citep{Raynor2019}. Indeed, the authors of the PadChest dataset bring up some of these caveats themselves, which are relevant even for the $27\%$ of their radiological reports that are text-mined by hand, which presumably have no \ac{NLP} errors~\cite{Aurelia_Bustos_2019}. An independent study of CXR14~\citep{Wang_2017} concludes that its labels have low positive predictive value and argues that visual inspection is necessary to create radiological datasets~\citep{Raynor2019}}.  Consequently, \ac{PLCO} is unique in that it is the only large-scale \ac{CXR} dataset with labels generated via \emph{visual observation} from radiologists. Although the \ac{PLCO} data is older than alternatives~\citep{Wang_2017,Aurelia_Bustos_2019,Irvin_2019,johnson_mimic-cxr_2019}, it has greater label reliability.

\begin{figure*}
\center
\includegraphics[scale = 0.2]{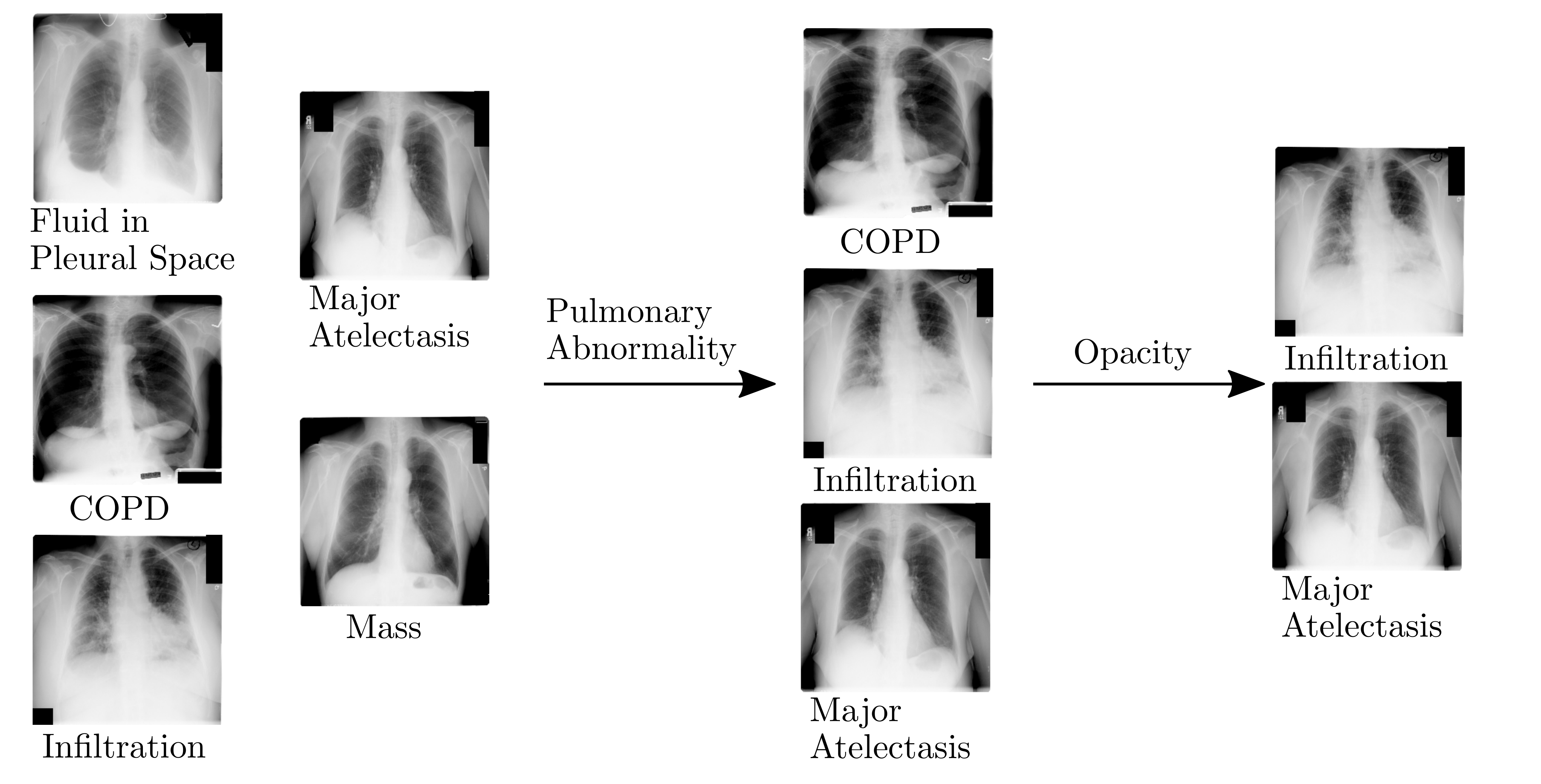}
\caption{Example \acs{PLCO} \acp{CXR} drawn from three levels of our taxonomy. On the left, at the higest level of taxonomy, \ie{} ``Abnormality'', disease patterns may manifest as a variety of visual features within the lung parenchyma, lung pleura, or the surrounding organs/tissues. As one progresses down the taxonomy, \ie{} to ``Opacity'', the discriminating task is narrowed into identifying the ``cloudy'' patterns seen in both ``Infiltration'' and ``Major Atelectasis.''}
\label{fig:visual_sample}
\end{figure*}

Radiologists in the \ac{PLCO} trial labelled $15$ disease patterns, which we call ``leaf labels'' in our taxonomy. Because of low prevalance, we merged ``left hilar abnormality'' and ``right hilar abnormality'' into ``hilar abnormality'', resulting in $14$ labels. From the leaf nodes, we constructed the label taxonomy
\begin{figure*}
\center
\includegraphics[scale = 0.13]{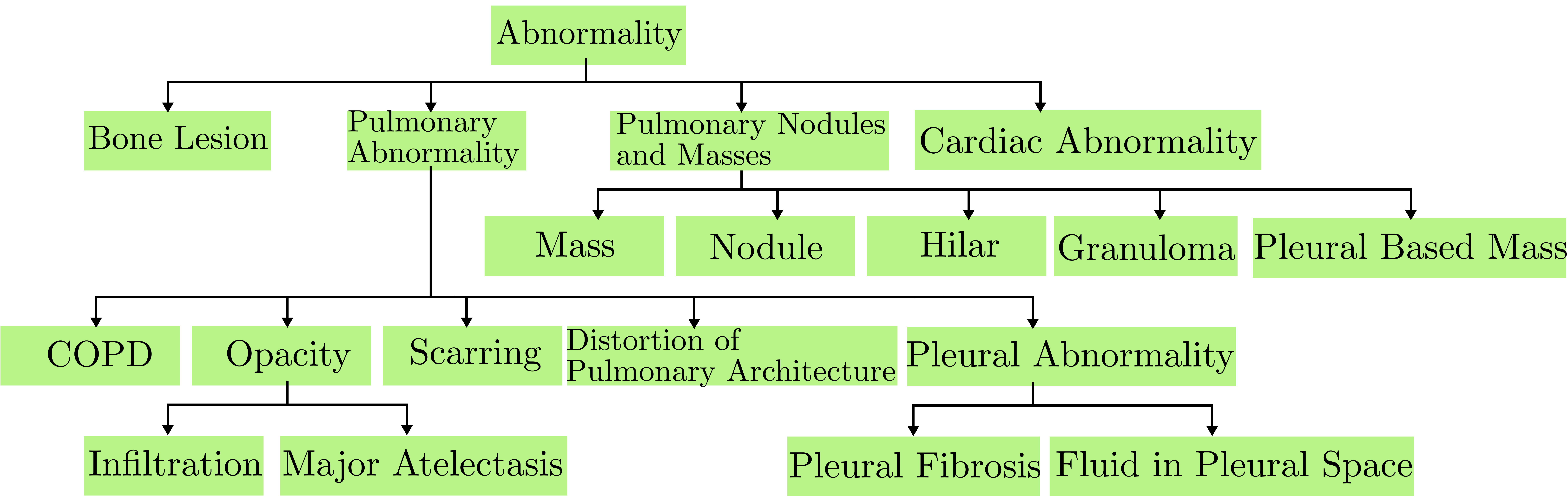}
\caption{Constructed label hierarchy from the \ac{PLCO} dataset.}
\label{fig:hierarchy}
\end{figure*}
shown in Figure~\ref{fig:hierarchy}. The hierarchical structure follows the \ac{PLCO} trial's division of ``suspicious for cancer'' disease patterns vs. not, and is further partitioned using common groupings~\citep{Folio_2012}, totalling $19$ leaf and non-leaf labels. While care was taken in constructing the taxonomy and we aimed for clinical usefulness, we make no specific claim as such. We instead use the taxonomy to explore the benefits of \ac{HMLC}, stressing that our approach is general enough to incorporate any appropriate taxonomy. \Fig~\ref{fig:visual_sample} visually depicts examples from our chosen \ac{CXR} taxonomy.

\begin{figure*}
\center
\includegraphics[scale = 0.12]{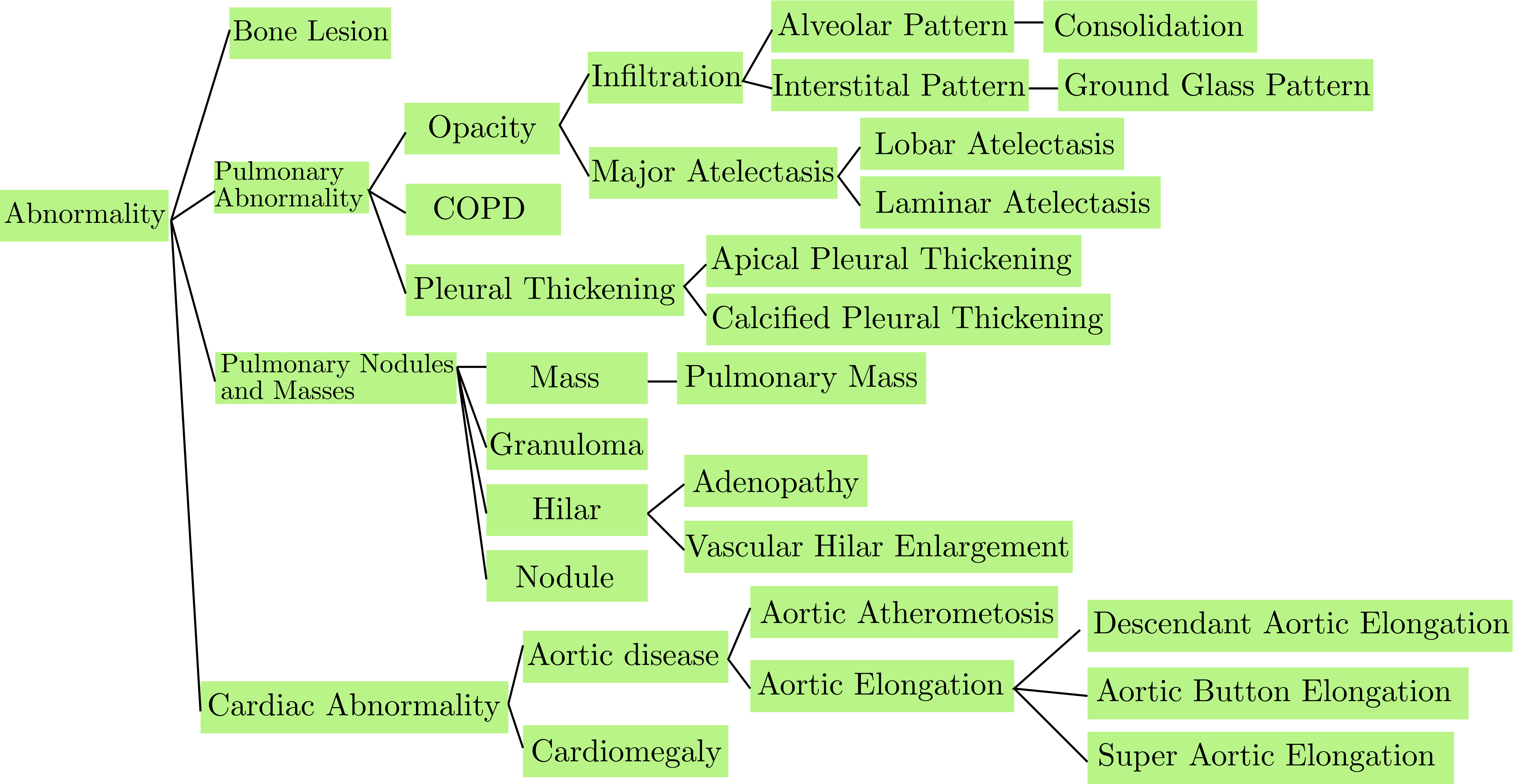}
\caption{Constructed label hierarchy from the {PadChest} dataset.}
\label{fig:hierarchical_label_structure_PadChest}
\end{figure*}

As supporting validation to our main \ac{PLCO} experiments, we also validate on the PadChest dataset~\cite{Aurelia_Bustos_2019}, which contains $160,845$ \acp{CXR} whose labels are drawn from either manual or automatic extraction from radiological \emph{text reports}. We focus on labels categorized as ``radiological findings'', which are more likely to correspond to actual disease patterns found on the \acp{CXR}~\cite{Aurelia_Bustos_2019}. Any \ac{CXR} with a solitary ``Unchanged'' label is removed, resulting in $121,242$ samples. Uniquely, PadChest offers a complete hierarchical structure for all labels. We remove labels with less than 100 \newtwo{manually labelled} samples and only retain labels that align with our \ac{PLCO} taxonomy. \newtwo{This both ensures we have enough statistical power for evaluation and that we are retaining \ac{PLCO}-like terms that we can confidently treat as clinically significant. As a result, total 30 out of 191 labels are selected, and our supplementary includes more details of the included and excluded labels.} The resulting taxonomy is shown in Figure~\ref{fig:hierarchical_label_structure_PadChest}. Unlike \ac{PLCO}, certain parent labels can be positive with no positive children labels, \eg{} ``Aortic Elongation''.

\subsection{Hierarchical Multi-Label Classification}
\label{sec:hmlc}

With a taxonomy established, a hierarchical approach to classification must be established. Because this is a multi-label setting, all or none of the labels in Figure~\ref{fig:hierarchy} can be positive. The only restriction is that if a child is positive, its parent must be too. Siblings are not mutually exclusive. For \ac{PLCO}, we assume that each image is associated with a set of ground-truth leaf labels and their antecedents, \ie{} there are no incomplete paths. However, for PadChest a ground-truth path may terminate before a leaf node. A training set, may have missing labels.

We use a DenseNet-121~\citep{Huang_2016} model as a backbone. If we use $k$ to denote the total number of leaf and non-leaf labels, we connect $k$ fully connected layers to the backbone's last feature layer to extract $k$ scalar outputs. Each output is assumed to represent the conditional probability (or its logit) given its parent is true. Thus, once the model is successfully trained, unconditional probabilities can be calculated from the output using the chain rule, \eg{} from the \ac{PLCO} taxonomy the unconditional probability of \textit{scarring} can be calculated as
\begin{align}
P(\text{Scar.}) = P(\text{Abn.})P(\text{Pulm.}|\text{Abn.})P(\text{Scar.}|\text{Pulm.})\mathrm{ ,}
\end{align}
where we use abbreviations for the sake of typesetting. In this way, the predicted unconditional probability of a parent label is guaranteed to be greater than or equal to its children labels. We refer to the conditional probability in a label hierarchy as \ac{HLCP}, and the unconditional probability calculated following the chain rule as \ac{HLUP}. The network outputs can be trained either conditionally or unconditionally, which we outline in the next two sections.

\subsection{Training with Conditional Probability}
\label{sec:conditional}

Similar to prior work~\citep{Redmon_2017,Roy_2018,Yan_2014}, in the first stage of the proposed training scheme, each classifier is only trained on data conditioned upon its parent label being positive. Thus, training directly models the conditional probability. The shared part of the classifiers, \ie{} feature layers from the backbone network, is trained jointly by all the tasks. Specifically, for each image the losses are only calculated on labels whose parent label is also positive. For example, and once again using the \ac{PLCO} taxonomy, when an image with positive \textit{Scarring} and no other positive labels is fed into training, only the losses of \textit{Abnormality} and the children labels of \textit{Pulmonary Abnormality} and \textit{Abnormality} are calculated and used for training. 
\begin{figure}
\center
\begin{tabular}{cc}
\includegraphics[scale=0.2]{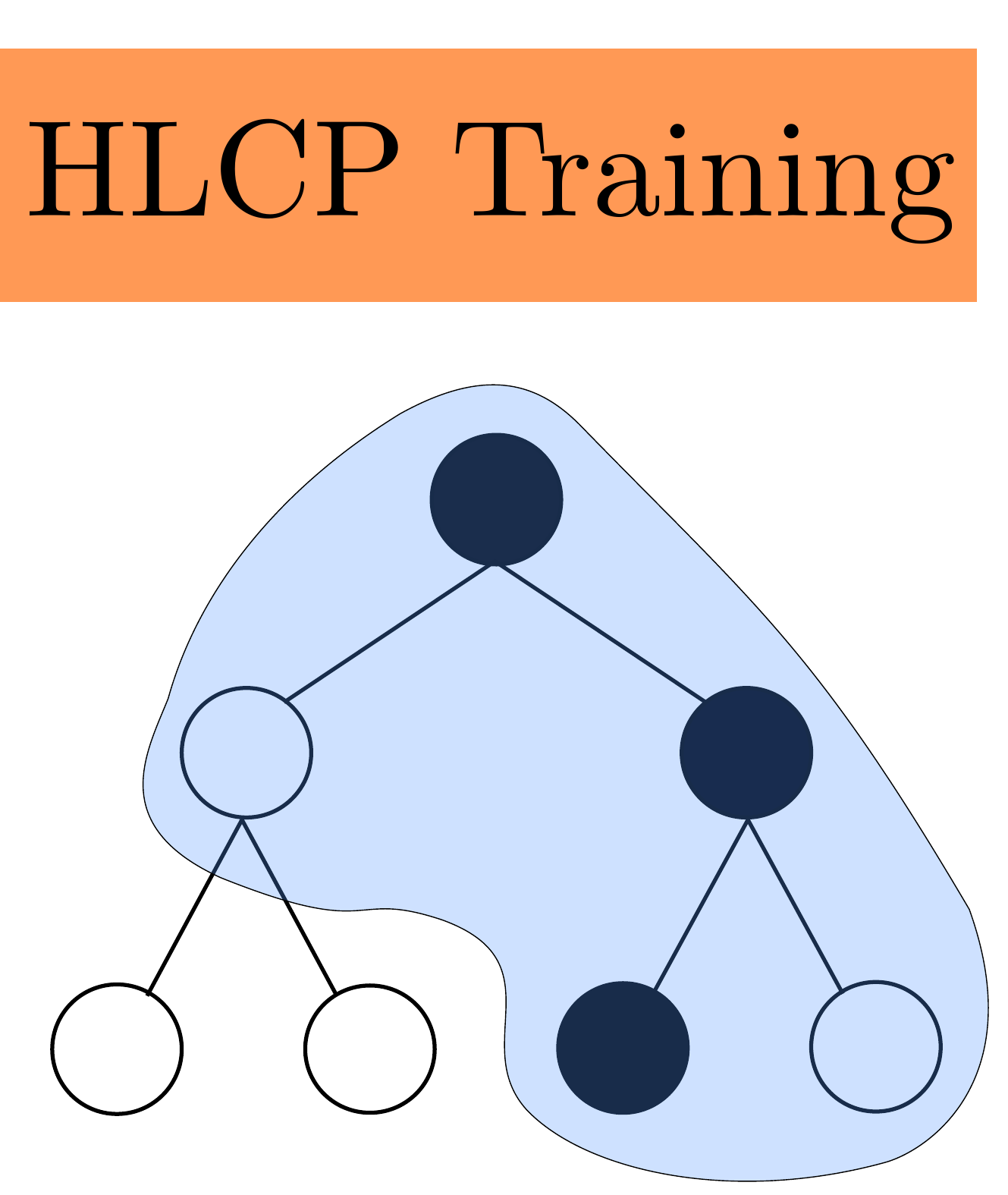} & \includegraphics[scale=0.2]{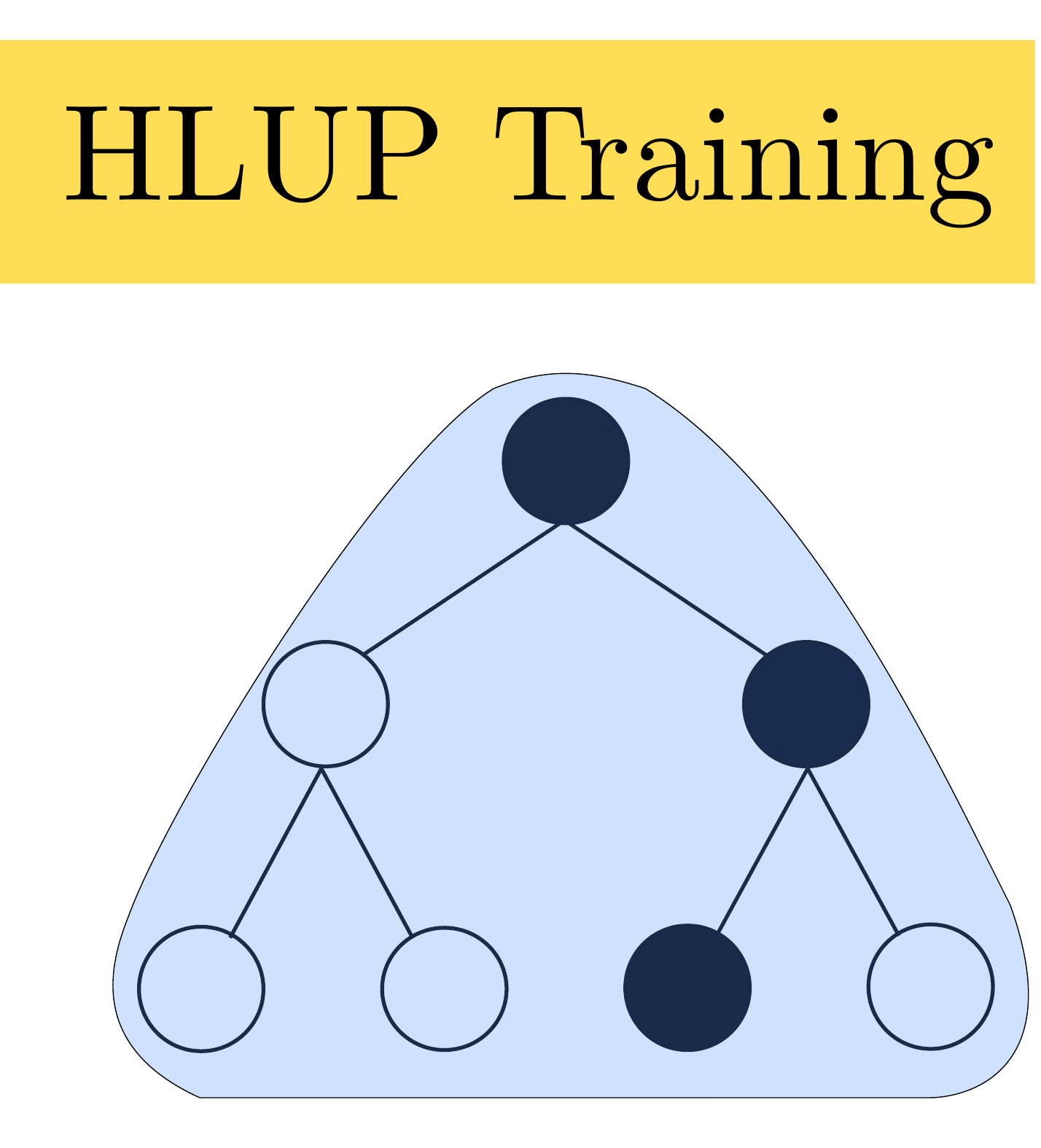} \\
(a) & (b)
\end{tabular}

\caption{The \ac{HLCP} and \ac{HLUP} losses are depicted in (a) and (b), respectively, where black and white points are positive and negative labels, respectively. Blue areas indicate the activation area in the loss functions.}
\label{fig:two_stages}
\end{figure}

Figure~\ref{fig:two_stages} (a) illustrates this training regimen, which we denote \ac{HLCP} training. In this work, we use \ac{CE} loss to train the conditional probabilities, which can be written as
\begin{align}
    L_{HLCP} = \sum_{m \in M} \mathit{CE}\left(z_{m}\mathrm{,}\, \hat{z}_m\right)*1_{\{z_{a(m)} =1\}}\mathrm{,} \label{eqn:hlcp_loss}
\end{align}
where $M$ denotes the set of all disease patterns, and $m$ and $a(m)$ denote a disease pattern and its ancestor, respectively. Here $\mathit{CE}(\cdot, \cdot)$ denotes the cross entropy loss, and $z_m \in \{0,1\}$ denotes the ground truth label of $m$, with $\hat{z}_{m}$ corresponding to the network's sigmoid output.

Training with conditional probability is a very effective initialization step, as it concentrates the modeling power solely on discriminating siblings under the same parent label, rather than having to discriminate across all labels, which eases convergence and reduces confounding factors. It also alleviates the problem of low label prevalence because fewer negative samples are used for each label.

\subsection{Fine Tuning with Unconditional Probability}
\label{sec:unconditional}

In the second stage, we finetune the model using an \ac{HLUP} \ac{CE} loss. This stage aims at improving the accuracy of unconditional probability predictions, which is what is actually used during inference and is thus critical to classification performance. Another important advantage is that the final linear layer sees more negative samples. Predicted unconditional probabilities for label $m$, denoted $\hat{p}_{m}$, are calculated using the chain rule:
\begin{align}
	\hat{p}_{m} = \prod_{m' \in A(m)}\hat{z}_{m'}\mathrm{,}
    \label{eqn:fullprob}
\end{align}
where $A(m)$ is the union of label $m$ and its antecedents. When training using unconditional probabilities, the loss is calculated on every classifier output for every data instance. Thus, the \ac{HLUP} \ac{CE} loss for each image is simply
\begin{align}
L_{HLUP} = \sum_{m \in M} \mathit{CE}\left(z_m\mathrm{,}\, \hat{p}_{m}\right)\mathrm{.} \label{eqn:hlup_loss}
\end{align}
Figure~\ref{fig:two_stages}(b) visually depicts this loss. 

A naive way to calculate \eqref{eqn:hlup_loss} would be a direct calculation. However, such an approach introduces instability during optimization, as the training would have to minimize the product of network outputs. In addition, the product of probability values within $[0,1]$ can cause arithmetic underflow. For this reason, we derive a numerically stable formulation below.

Denoting the network's output logits as $\hat{y}_{(.)}$, the predicted unconditional probability of label $m$ can be written as:
\begin{align}
\hat{p}_m = \prod_{m'}\dfrac{1}{1+\exp(-y_{m'})},
\end{align}
where we use $m'$ to denote $m'\in A(m)$ for notational simplicity.

The \ac{HLUP} \ac{CE} loss is calculated as:
\begin{align}
	L_{HLUP} =& -z_m\log(\hat{p}_{m}) -(1-z_m)\log(1-\hat{p}_{m}) \mathrm{,} \\
	=& -z_m\log\left(\prod_{m'} \dfrac{1}{1+\exp(-y_{m'})}\right) \nonumber\\
	&-(1-z_m)\log \left(1-\left(\prod_{m'} \dfrac{1}{1+\exp(-y_{m'})}\right)\right) \mathrm{,} 
\label{eqn:raw}
\end{align}
where $z_m$ is the ground truth label of $m$. 

The formulation in \eqref{eqn:raw} closely resembles several cross-entropy loss terms combined together. To see this, we can break up the second term in \eqref{eqn:raw} to produce the following formulation:
\begin{align}
\label{eqn:stable_hlup}
L_{HLUP} =&-z_m \log\left(\prod_{m'}\dfrac{1}{1+\exp(-y_{m'})}\right) \nonumber\\ &-(1-z_m)\log\left(\prod_{m'}\left(1-\dfrac{1}{1+\exp(-y_{m'})}\right)\right) + \gamma \mathrm{,}
\end{align} 
where $\gamma$ is a scalar quantity that must be formulated. The log terms above can then be decomposed as
\begin{align}
L_{HLUP}=& \sum_{m'} \left( -z_m \log\left(\dfrac{1}{1+\exp(-y_{m'})}\right)\right. \nonumber\\ & \left.-(1-z_m)\log\left(1-\dfrac{1}{1+\exp(-y_{m'})}\right) \right)+ \gamma \mathrm{,} \\
 =& \sum_{m'}\ell_{m'} +\gamma \textrm{,} 
\label{eqn:better_raw}
\end{align}
where $\ell_{m}$ are individual cross entropy terms, using $z_m$ and $y_{m'}$ as the ground truth and logit input, respectively. Note that \eqref{eqn:better_raw} allows us to take advantage of numerically stable \ac{CE} implementations to calculate $\sum_{m'}\ell_{m'}$. However to satisfy \eqref{eqn:better_raw}, we will need $\gamma$ to satisfy:
\begin{align}
 \gamma =& (1-z_m)\log\left(\prod_{m'}\left(1-\dfrac{1}{1+\exp(-y_{m'})}\right)\right) \nonumber \\
 &-(1-z_m)\log \left(1-\left(\prod_{m'} \dfrac{1}{1+\exp(-y_{m'})}\right)\right)\mathrm{,}  \\
  =& (1-z_m)\log\left(\dfrac{\prod_{m' }\exp(-y_{m'})}{\prod_{m' }(1+\exp(-y_{m'}))}\right) \nonumber  \\
 &-(1-z_m)\log\left( \dfrac{\prod_{m'}(1+\exp(-y_{m'})) -1 }{\prod_{m'}(1+\exp(-y_{m'}))}\right) \mathrm{,} \\
 =&(1-z_m)\log\left(\dfrac{\exp(\sum_{m'}-y_{m'})}{\prod_{m' }(1+\exp(-y_{m'})) -1}\right) \mathrm{,}\\
 =&(1-z_m)\left(\sum_{m'}-y_{m'}-\log\left(\prod_{m'}(1+\exp(-y_{m'})) -1\right)\right) \label{eqn:gamma_pre} \textrm{.} 
\end{align}

If the product within the log-term of \eqref{eqn:gamma_pre} is expanded, with $1$ subtracted, it will result in
\begin{align}
	\gamma=(1-z_m)\left(\sum_{m'}-y_{m'}-\log\left( \sum_{S \in \mathcal{P}(A(m))\setminus\{\emptyset\}} \exp\left(\sum_{j\in S}-y_{j}\right)\right)\right) \textrm{,} \label{eqn:gamma_expanded}
\end{align}
where $S$ enumerates all possible subsets of the powerset of $A(m)$, excluding the empty set. For example if there were two logits, $y_{1}$ and $y_{2}$, the summation inside the log would be:
\begin{align}
	\exp(-y_{1}) + \exp(-y_{2}) + \exp(-y_{1}-y_{2}) \textrm{.}
\end{align}
The expression in \eqref{eqn:gamma_expanded} can be written as
\begin{align}
	\gamma=(1-z_m)\left(\sum_{m'}-y_{m'}-\mathit{LSE}\left(\left\{\sum_{j\in S}-y_j \quad \forall S \in \mathcal{P}(A(m))\setminus\{\emptyset\} \right \}\right)\right) \textrm{,} \label{eqn:final_gamma_app}
\end{align}
where $\mathit{LSE}$ is the LogSumExp function. Numerically stable implementations of the LogSumExp, and its gradient, are well known. By substituting \eqref{eqn:final_gamma_app} into \eqref{eqn:better_raw}, a numerically stable version of the \ac{HLUP} \ac{CE} loss can be calculated. 

Enumerating the powerset produces an obvious combinatorial explosion. However, for smaller-scale hierarchies, like that in Figure~\ref{fig:hierarchy}, it remains tractable. For larger hierarchies, an $O(|A(m)|)$ solution involves simply interpreting the LogSumExp as a smooth approximation to the maximum function, which we provide here for completeness:
\begin{align}
	\gamma &\approx(1-z_m)\left(\sum_{m'}-y_{m'}-\max\left(\left\{\sum_{j\in S}-y_j \quad \forall S \in \mathcal{P}(A(m))\setminus\{\emptyset\} \right \}\right)\right) \textrm{,} \\
	&=\begin{cases}
	(1-z_m)\left(\sum_{m'}-y_{m'}-\sum_{j: y_{j}<0}{-y_{j}}\right) \textrm{,} & \text{if } \exists \, y_{m'} < 0 \\
	(1-z_m)\left(\sum_{m'}-y_{m'} -\max(\{-y_{m'}\})\right) \textrm{,} & \text{otherwise}
	\end{cases} \textrm{.}
\end{align}


\section{Experimental}

We perform two types of experiments to validate our \ac{HMLC} approach. The first uses the standard completely labelled setup, helping to reveal how our use of taxonomic classification can help produce better raw classification performance than typical ``flat'' classifiers. The second uses incompletely labelled data under controlled scenarios to show how our \ac{HMLC} approach can naturally handle such data, achieving even higher boosts in relative performance.

\subsection{Complete Labels}
\label{sec:complete_experimental}

\noindent \textbf{Experimental Setup}
We test our \ac{HMLC} approach on both the \ac{PLCO}~\cite{Gohagan_2000} and PadChest~\cite{Aurelia_Bustos_2019} datasets, using the taxonomies of \Fig~\ref{fig:hierarchy} and \Fig~\ref{fig:hierarchical_label_structure_PadChest}, respectively. Our emphasis is on \ac{PLCO} due to its more reliable labels, but evaluations on PadChest provide important experimental support, especially given its larger taxonomy. Following accepted practices in large-scale \ac{CXR} classification~\cite{Wang_2017,Irvin_2019,Aurelia_Bustos_2019}, we split the data into single training, validation, and test sets, corresponding to $70\%$, $10\%$, and $20\%$ of the data, respectively. Data is split at the patient level, and care was taken to balance the prevalence of each disease pattern as much as possible. As mentioned above, our \ac{HMLC} approach uses a trunk network, with a final fully-connected layer outputting logit values for each of the nodes of our chosen taxonomy. Our chosen network is DenseNet-121~\citep{Huang_2016}, implemented using TensorFlow. We first train with the \ac{HLCP} \ac{CE} loss of \eqref{eqn:hlcp_loss} fine-tuning from a model pretrained from ImageNet~\citep{imagenet_cvpr09}.  We refer to this model simply as \textit{\acs{HLCP}}. To produce our final model, we then finetune the \ac{HLCP} model using the \acs{HLUP} \ac{CE} loss of \eqref{eqn:hlup_loss}. We denote this final model as \textit{\ac{HLUP}-finetune}. 

\noindent \textbf{Comparisons}
In addition to comparing against \acs{HLCP}, we also compare against three other baseline models, all using the same trunk network fine-tuned from ImageNet pretrained weights. The first, denoted \textit{\acs{BR}-leaf}, is trained using \ac{CE} loss on the $14$ fine-grained labels. This measures performance using a standard multi-label \ac{BR} approach. The second, denoted \textit{\acs{BR}-all} is very similar, but trains a \ac{CE} loss on all labels independently, including non-leaf ones. In this way, \textit{\ac{BR}-all} measures performance when one wishes to naively output non-leaf abnormality nodes, without considering label taxonomy. Finally, we also test against a model trained using the \ac{HLUP} \ac{CE} loss directly from ImageNet weights, rather than finetuning from the \ac{HLCP} model. As such, this baseline, denoted \textit{\acs{HLUP}}, helps reveal the impact of using a two-stage approach vs. simply training an \ac{HLUP} classifier in one step. For all tested models, extensive hyper-parameter searches were performed on the NVIDIA cluster to optimize mean validation fine-grained \acp{AUC}. 

For comparisons to external models, we also compare to a recent DenseNet121 \ac{BR} approach~\citep{guendel_2019} trained on the \ac{PLCO} data. But, we stress that direct comparisons of numbers are impossible, as \citep{guendel_2019} used different data splits and only evaluated on $12$ fine-grained labels. In the interest of fairness we compare against both (a) their best reported numbers when only training a classifier on \ac{CXR} disease patterns and (b) their best reported numbers overall, in which the authors incorporated segmentation and localization cues. For (a), we use numbers reported on an earlier work~\citep{guendel2018learning}, which were higher. Unfortunately, both sets of their reported numbers are based on training data that also included the ChestXRay14 dataset~\citep{wang_chestnet:_2018}, providing an additional confounding factor that hampers any direct comparison.

Finally, we also run experiments to compare our numerically stable implementation of \ac{HLUP} \ac{CE} loss in \eqref{eqn:stable_hlup} to: (a) the naive approach of directly optimizing \eqref{eqn:fullprob}; and (b) to a recent rescaling approximation, originally introduced for the multiplication of independent, rather than conditional probabilities, seen in multi-instance learning~\citep{li_thoracic_2018}. This latter approach re-scales each individual probability multiplicand (term) in \eqref{eqn:fullprob} to guarantee that the product is greater than or equal to $1\mathrm{e}\textnormal{-}7$. Similar to the naive approach, the product is then optimized directly using \ac{CE} loss. For the \ac{PLCO} dataset, based on a maximum depth of four for the taxonomy, we implement this approach by re-scaling each multiplicand in \eqref{eqn:fullprob} to $[0.02,1]$. 


\noindent \textbf{Evaluation Metrics}
We evaluate our approach using \ac{AUC} and \ac{AP}, calculated across both leaf and non-leaf labels, when applicable. Additionally, we also evaluate using  conditional \ac{AUC} and \ac{AP} metrics, which are metrics that reflect the complicated evaluation space of multi-label classification. In short, because more than one label can be positive, multi-label classification performance has exponentially more facets for evaluation than single-label or even multi-class settings. Conditional metrics are one such facet, that focus on model performance conditioned on certain non-leaf labels being positive. Here, we restrict our focus to \acp{CXR} exhibiting one or more disease patterns, \ie{} \textit{abnormality} being positive. As such, this sheds light on model performance when it may be critical to discriminate what combination of disease patterns are present, which is crucial for proper \ac{CXR} interpretation~\citep{Folio_2012}. 

\subsection{Incomplete Labels}

\noindent \textbf{Experimental Setup}
We also use the \ac{PLCO} dataset~\citep{Gohagan_2000} to characterize the benefits of our \ac{HMLC} approach when faced with incomplete labels. However, after publication of our original work~\citep{chen_deep_2019}, the \ac{PLCO} organizers altered their data release policies and only released a subset of the original dataset, containing $88\,737$ labeled \acp{CXR} from $24\,997$ patients\footnote{The first author no longer had access to the original dataset for the incomplete label experiments as he had finished his internship at NVIDIA.}. For this reason, we perform our incomplete labels experiments on this smaller dataset, splitting and preparing the data in an identical manner as described in \Sec~\ref{sec:complete_experimental}. 

To simulate a scenario where learning algorithms may be faced with incomplete labels, we removed known labels from the training set using the following controlled scheme:
\begin{enumerate}
    \item We choose a base deletion probability, $\beta\in[0,1]$.
    \item For data instances with positive labels for ``Pleural Abnormality'', ``Opacity'', and ``Pulmonary Nodules and Masses'', we delete all their children labels with a probability of $\beta$. For example, if we delete the children labels of a positive ``Pleural Abnormality'' instance, then it is no longer known whether the ``Pleural Abnormality'' label corresponds to  ``Pleural Fibrosis'', or ``Fluid in Pleural Space'', or both.
    \item We perform the same steps for data instances with positive labels for ``Pulmonary Abnormality'' and ``Abnormality'', except with probabilities of $0.3\beta$ and $0.3^2\beta$, respectively. For example, if the children of a positive instance of ``Abnormality'' were deleted, then it is only known there are one or more disease patterns present, but not which one(s). 
    \item A higher-level deletion overrides any decision(s) at finer levels. 
    \item Because of their extremely low prevalence, we ignore the ``Major Atelectasis'' and ``Distortion in Pulmonary Architecture'' labels in training and evaluation.
\end{enumerate}
Note that this scheme makes it more likely to have a missing fine-grained label over a higher-level label, which we posit follows most scenarios producing incomplete labels.  When labels are deleted, we treat them as unknown and do not execute any training loss on them. We test our \ac{HMLC} algorithm and baselines on the following $\beta$ values: $\{0,.1,.2,.3,.4,.5,.6,.7\}$, which ranges from no incompleteness to roughly $70\%$ of fine-grained labels being deleted. To allow for stable comparisons across $\beta$ values, we also ensure that if a label was deleted at a certain value of $\beta$, it will also be deleted at all higher values of $\beta$. To ease reproducibility, we publicly release our data splits\footnote{https://github.com/hchen135/Hierarchical-Multi-Label-Classification-X-Rays}. All other implementation details are also identical to that of \Sec~\ref{sec:complete_experimental}.

\noindent \textbf{Evaluation Metrics and Comparisons}
We measure \ac{AUC} values and compare our chosen model of \ac{HLUP} finetune against \ac{BR}-leaf and \ac{BR}-all. 

\section{Results and Discussion}

We focus in turn on experiments with complete and incomplete labels, which can be found in \Sec~\ref{sec:complete} and \Sec~\ref{sec:incomplete}, respectively.

\subsection{Complete Labels}
\label{sec:complete}
Our complete labels experiments first focus on the benefits of our \ac{HLUP}-finetune approach compared to alternative ``flat'' and \ac{HMLC} strategies. Then, we discuss results specifically focusing on our numerically stable \ac{HLUP} \ac{CE} loss.



\subsubsection{HLUP-finetune Performance}

Table~\ref{tab:AUCAP} outlines the \ac{PLCO} results of our \acs{HLUP}-finetune approach vs. competitors.
\begin{table*}[t]
\small
\begin{center}
	\caption{\emph{PLCO} \ac{AUC} and \ac{AP} values across tested models. Mean values across  leaf and non-leaf disease patterns are shown, as well as for leaf labels conditioned on one or more abnormalities being present.}
    \label{tab:AUCAP}
	 \begin{tabular}{l|cc|cc|cc}	
	 	\toprule
	 	~ & \multicolumn{2}{C{.16\textwidth}|}{\textbf{Leaf labels}} & \multicolumn{2}{C{.16\textwidth}|}{\textbf{Non-leaf labels}} & \multicolumn{2}{C{.35\textwidth}}{\textbf{Leaf labels conditioned on abnormality}}\\
	 	\hline
	 	~ 				&   AUC	& AP  		&AUC		&  AP	&AUC		&AP\\
	 	\hline
	 	\citep{guendel2018learning} & $0.865$ &N/A & N/A & N/A & N/A & N/A  \\
 	 	\hline
 		\citep{guendel_2019}			& $0.883$ 				& N/A 					&    N/A 	 				&N/A	&	N/A & N/A		\\
	 	\hline
		\ac{BR}-leaf 			&$0.871$ 				&$0.234$ 					&    N/A 	 				&N/A	&	$0.806$ & $0.334$		\\
		
	 	\hline
	 	\ac{BR}-all			&$0.867$       				&$0.221$					&$0.852$      				&$0.440$	& $0.808$ & $0.323$		\\
	 	\hline
    	HLUP 		&$0.872$       		 		&$0.214$  					&   $0.856$ 				&$0.436$	& $0.799$ & $0.288$		\\
	 	\hline
		HLCP			&$0.879$           			&$0.229$ 					&$0.857$				&$0.440$	& $0.822$	& $0.329$	\\
	 	\hline
	 	HLUP-finetune		&$\boldsymbol{0.887}$        			&$\boldsymbol{0.250}$ 					& $\boldsymbol{0.866}$    				&$\boldsymbol{0.460}$	& $\boldsymbol{0.832}$ & $\boldsymbol{0.342}$		\\
	 	\bottomrule
	 \end{tabular}
\end{center}
\end{table*}
As the table demonstrates, the standard baseline \ac{BR}-leaf model produces high \ac{AUC} scores, in line with prior work~\citep{guendel2018learning}; however, it does not provide high-level predictions based on a taxonomy. Naively executing \ac{BR} training on the entire taxonomy, \ie{} the \ac{BR}-all model, does not improve performance. This indicates that if not properly incorporated, the label taxonomy does not benefit performance.

In contrast, the \ac{HLCP} model is indeed able to match \ac{BR}-leaf's performance on the fine-grained labels, despite also being able to provide high-level predictions. \ac{HLUP}-finetune goes further by exceeding \ac{BR}-leaf's fine-grained performance, demonstrating that our two-stage training process can produce tangible improvements. This is underscored when comparing \ac{HLUP}-finetune with \ac{HLUP}, which highlights that without the two-stage training, \ac{HLUP} training cannot reach the same performance. If we limit ourselves to models incorporating the entire taxonomy, our final \ac{HLUP}-finetune model outperforms \ac{BR}-all by $2\%$ and $2.9\%$ in leaf-label mean \ac{AUC} and \ac{AP} values, respectively. Because \ac{HLUP}-finetune shares the same labels as \ac{BR}-all, the performance boosts of the former over the latter demonstrate that the additional output nodes seen in \ac{HMLC} are not responsible for performance increases. Instead, it is indeed the explicit incorporation of taxonomic structure that leads to improved performance. 

\begin{figure*}[t]
\center
\includegraphics[scale = 0.55]{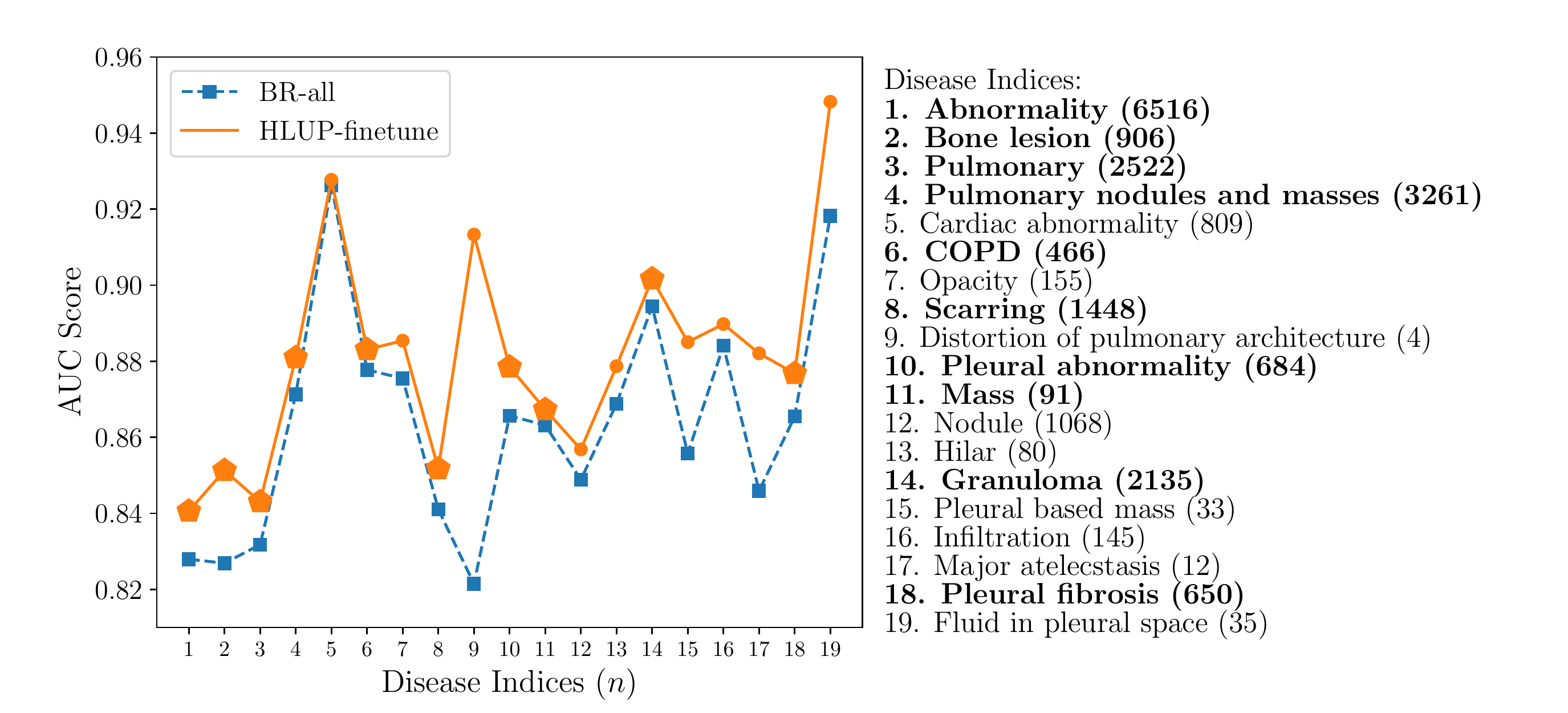}
\caption{Comparison of \ac{AUC} scores for all fine-grained and high-level (non-leaf) disease patterns for the \ac{BR}-all and \ac{HLUP}-finetune models. The dashed line separates the fine-grained from the high-level (non-leaf) disease patterns. Boldface labels and larger graph markers denote disease patterns exhibiting statistically significant improvement ($p<0.05$) using the StAR software implementation~\citep{Vergara_2008} of the non-parametric test of \citep{DeLong_1988}.}
\label{fig:auc}
\end{figure*}
Figure~\ref{fig:auc} provides more details on these improvements, demonstrating that \ac{AUC} values are higher for \ac{HLUP}-finetune compared to the baseline method for all fine-grained and high-level disease patterns. Interested readers can find these \ac{AUC} values in our supplementary materials. Although not graphed here for clarity reasons, \ac{HLUP}-finetune also outperformed the \ac{HLCP} method for all disease patterns. Of note is that statistically significant differences also respect the disease hierarchy, and if a child disease pattern demonstrates statistically significant improvement, so does its parent.

Of particular note, when considering \acp{AUC} conditioned on one or more abnormalities being present (last column of Table~\ref{tab:AUCAP}), the gap between all \ac{HMLC} approaches and ``flat'' classifiers increases even more. As can be seen in such settings, \ac{HLUP}-finetune still exhibits increased performance over the baseline models and also the next-best hierarchical model. Importantly, if we compare the conditional \acp{AUC} between \ac{BR}-all and \ac{HLUP}-finetune, we see a $2.4\%$ increase. This indicates that \ac{HMLC} is particularly effective at differentiating the exact combination of abnormalities present within an image.  This may reduce the amount of spurious and distracting predictions upon deployment, but more investigation is required to quantify this.  

We also note that \ac{HLUP}-finetune managed to outperform \citep{guendel_2019}'s \ac{AUC} numbers,  despite the latter incorporating almost twice the amount of data and also including additional localization and segmentation tasks. However, we again note that \citep{guendel_2019} used a different data split and only $12$ fine-grained labels, so such comparisons can only be taken so far.

\begin{table*}[t]
\small
\begin{center}
	\caption{ \emph{PadChest} \ac{AUC} and \ac{AP} values across tested models. Mean values across  leaf and non-leaf disease patterns are shown, as well as for leaf labels conditioned on one or more abnormalities being present.}
    \label{tab:AUCAP_padchest}
	 \begin{tabular}{l|cc|cc|cc}	
	 	\toprule
	 	~ & \multicolumn{2}{C{.16\textwidth}|}{\textbf{Leaf labels}} & \multicolumn{2}{C{.16\textwidth}|}{\textbf{Non-leaf labels}} & \multicolumn{2}{C{.35\textwidth}}{\textbf{Leaf labels conditioned on abnormality}}\\
	 	\toprule
	 	~ 				&   AUC	& AP  		&AUC		&  AP	&AUC		&AP\\
	 	\midrule
		\ac{BR}-leaf 			&$0.825$ 				&$0.104$ 					&    N/A 	 				&N/A	&	$0.743$ & $0.212$		\\
		
	 	\hline
	 	\ac{BR}-all			&$0.825$       				&$0.110$					&$0.820$      				&$0.221$	& $0.739$ & $0.204$		\\
	 	\hline
    	HLUP 		&$0.831$       		 		&$0.114$  					&   $0.828$ 				&$0.220$	& $0.752$ & $0.211$		\\
	 	\hline
		HLCP			&$0.831$           			&$0.135$ 					&$0.833$				&$0.240$	& $0.765$	& $0.244$	\\
	 	\hline
	 	HLUP-finetune		&$\boldsymbol{0.837}$        			&$\boldsymbol{0.145}$ 					& $\boldsymbol{0.840}$    				&$\boldsymbol{0.253}$	& $\boldsymbol{0.778}$ & $\boldsymbol{0.261}$		\\
	 	\bottomrule
	 \end{tabular}
\end{center}
\end{table*}
Experiments on PadChest further support these results, with trends mirroring that of the \ac{PLCO} experiments. As can be seen in Table~\ref{tab:AUCAP_padchest}, HLUP-finetune outperforms both the \ac{BR} baselines and \ac{HMLC} alternatives. Moreover, just like the \ac{PLCO} experiments, when evaluating \ac{AUC} and \ac{AP} conditioned on one or more abnormalities being present, the performance gaps between HLUP-finetune and alternatives further increase. The relative performance improvements demonstrate that our \ac{HMLC} approach generalizes well to a different \ac{CXR} dataset outside of \ac{PLCO}, even though PadChest uses a different taxonomy and was collected with very different patient populations at a much later date.

The \ac{PLCO} and PadChest performance boosts are in line with prior work that reported improved classification performance when exploiting taxonomy, \eg{} for text classification~\citep{mccallum_improving_1998,dumais_hierarchical_2000}, but here we use \ac{HMLC} in a more modern deep-learning setting and for an imaging-based \ac{CAD} application. In particular, given that taxonomy and ontology are crucial within medicine, the use of hierarchy is natural. Because the algorithmic approach we take remains very simple, our \ac{HMLC} approach may be an effective method for many other medical classification tasks outside of \acp{CXR}. 

The discussion of the performance boosts garnered by \ac{HMLC} are very important, but it should also be noted that \ac{HMLC} provides inherent benefits outside of raw classification performance. By ensuring that clinical taxonomy is respected, \ie{} a parent label's pseudo-probability will always be greater than or equal to any of its children's, \ac{HMLC} provides a more interpretable and understandable set of predictions that better match the top-down structure of medical ontology.

In addition to exploring the benefits of the conceptual approach of \ac{HMLC} to \ac{CXR} classification,  our work also demonstrates that a two-stage \ac{HLUP} finetuning approach can provide performance boosts over the more common one-stage \ac{HLCP} training seen in many prior deep-learning works~\citep{Redmon_2017,Roy_2018, Yan_2014}. As such, our two-stage approach may also prove useful to hierarchical classifiers seen in other domains, such as computer vision or text classification.

\subsubsection{Numerically Stable \ac{HLUP}}

\begin{table*}
\small
\begin{center}
	\caption{Comparison of \acsp{AUC} produced using different \ac{HLUP} \ac{CE} loss implementations for \ac{PLCO}. }
	\label{tab:google}
	\begin{tabular}{C{.12\textwidth}C{.12\textwidth}C{.12\textwidth}C{.12\textwidth}C{.12\textwidth}C{.12\textwidth}}
	 \toprule
	    HLUP (naive)  & HLUP (rescale) & HLUP (ours) & HLUP-finetune (naive) & HLUP-finetune (rescale) & HLUP-finetune (ours) \\
	    \midrule
	    $0.864$ & $0.853$ & $0.872$ & $ 0.886$ & $0.867$ & $0.887 $ \\
	     \bottomrule
	\end{tabular}
\end{center}
\end{table*}
\Tab~\ref{tab:google} demonstrates that our numerically stable \ac{HLUP} \ac{CE} loss results in much better \acp{AUC} compared to the competitor rescaling approach~\citep{li_thoracic_2018} and to naive \ac{HLUP} training when starting from ImageNet weights. However, there were no performance improvements when compared to the naive approach when finetuning from the \ac{HLCP} weights. We hypothesize that the predictions for the \ac{HLCP} are already at a sufficient quality that the numerical instabilities of the naive \ac{HLUP} \ac{CE} loss are not severe enough to impair performance. Nonetheless, given the improvements when training from ImageNet weights, these results indicate that our \ac{HLCP} \ac{CE} loss does indeed provide tangible improvements in convergence stability. We expect these improvements to be greater given taxonomies of greater depth, and our formulation should also prove valuable to multi-instance setups which must optimize \ac{CE} loss over the product of large numbers of probabilities, \eg{} the $256$ multiplicands seen in \citep{li_thoracic_2018}.

\subsection{Incomplete Labels}
\label{sec:incomplete}

\begin{figure}[t]
\center
\includegraphics[scale=0.5]{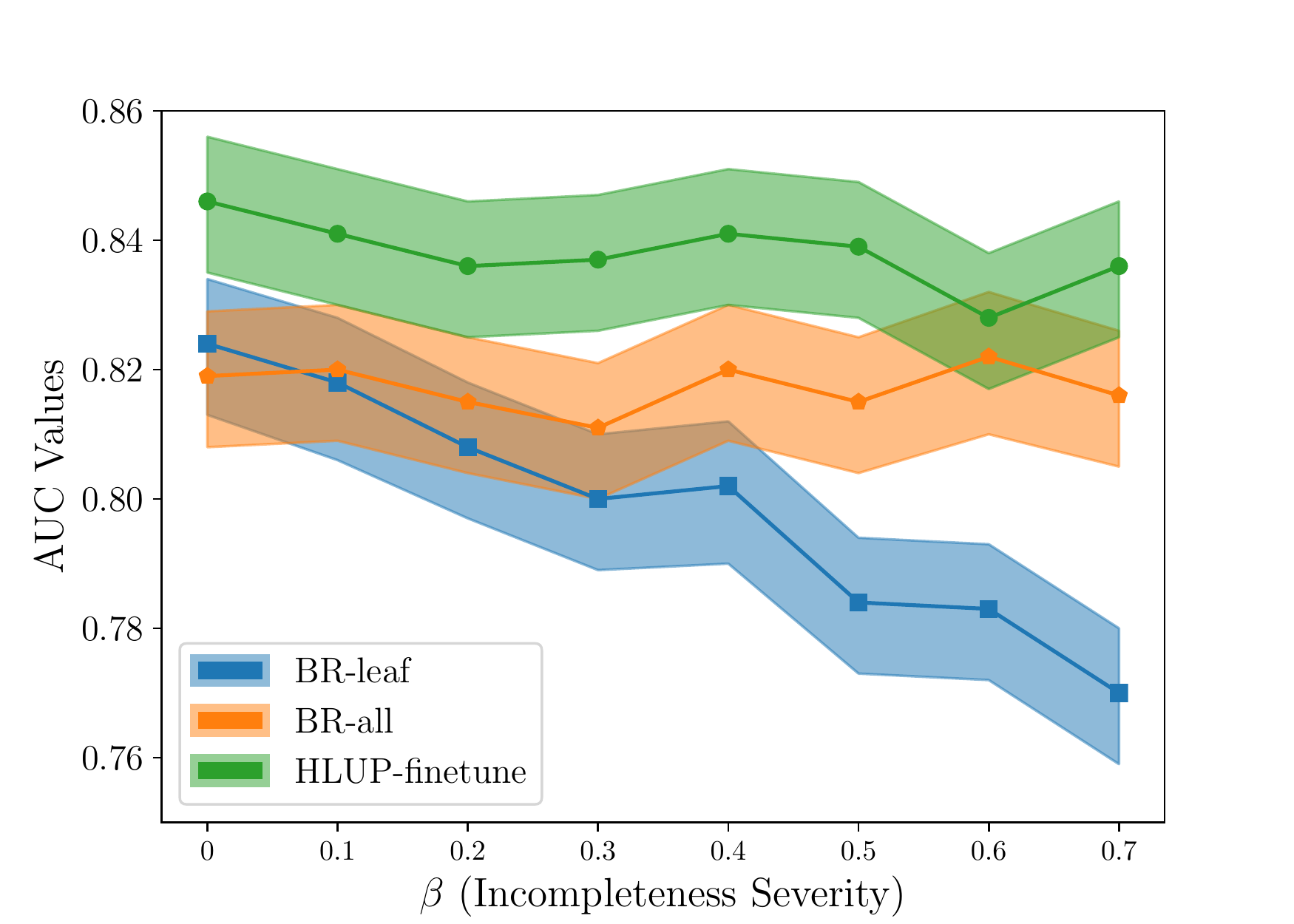}

\caption{Mean \ac{AUC} scores under different levels of label incompleteness with confidence intervals representing the 2.5th and 97.5th percentiles of $5000$ resampling with replacement bootstrap rounds~\citep{Dekking_2005}.}
\label{fig:incomplete}
\end{figure}

\Fig~\ref{fig:incomplete} shows the results of our incompletely labelled experiments. As can be seen when all labels are present, \ie{} $\beta=0$, the results mirror that of \Sec~\ref{sec:complete}, with \ac{HLUP}-finetune outperforming the baseline models and the BR-all providing no improvements over BR-leaf. As the incompleteness severity increases, \ac{BR}-leaf's performance drastically drops, while \ac{BR}-all and \ac{HLUP}-finetune are much better able to manage label incompleteness. At the highest $\beta$ level, the performance gap between \ac{HLUP}-finetune and \ac{BR}-leaf almost reaches $7\%$.  Per-abnormality \ac{AUC} values can be found in our supplementary materials. 

Our results demonstrate that incorporating hierarchy can be an effective means to manage incomplete labels. Specifically, while \ac{HLUP}-finetune's performance does indeed drop as the incompleteness severity increases, it does so at a drastically reduced rate compared to the standard BR-leaf classifier. Interestingly, BR-all, which trains all outputs but without incorporating a taxonomy, also manages to retain an equally graceful performance drop. However, \ac{HLUP}-finetune's roughly $2\%$ \ac{AUC} performance advantage over BR-all indicates that properly incorporating the taxonomic hierarchy is necessary to boost classification performance. We suspect the anomaly at $\beta=0.6$ is due to variability caused by the randomness of the training procedure and we reran our experiments at this $\beta$ value which confirmed this. Ideally, running multiple training runs at each $\beta$ value would allow us to produce confidence bars that take into account effects from random weight initialization and sampling, but time and computational resources did not allow us to perform this extremely demanding set of experiments. Finally, \ac{HLUP}-finetune has the added important benefit of producing predictions that respect the taxonomy, which is something that \ac{BR}-all does not do. Thus, these results indicate that when possible, incorporating a \ac{HMLC} approach can be an effective means to manage incompletely labelled data. As the prevalence of text-mined \ac{PACS} medical imaging data increases, we expect the need for approaches to gracefully handle missing labels to increase, and our \ac{HMLC} approach may provide a useful cornerstore of future work in this direction.

\section{Conclusions}

We have presented a two-stage approach for deep \ac{HMLC} of \acp{CXR} that combines conditional training with an unconditional probability fine-tuning step. To effect the latter, we introduce a new and numerically stable formulation for \ac{HLUP} \ac{CE} loss, which we expect would also prove valuable in other training scenarios involving the multiplication of probability predictions, \eg{} multi-instance learning. Through comprehensive evaluations, we report the highest mean \ac{AUC} on the \ac{PLCO} dataset yet, outperforming hierarchical and non-hierarchical alternatives. Supporting experiments on the PadChest dataset confirm these results. We also show performance improvements conditioned on one or more abnormalities being present, \ie{} predicting the specific combination of disease patterns, which is crucial for \ac{CXR} interpretation. Experiments with incompletely labelled data also demonstrate that our two-stage \ac{HMLC} approach is an effective means to handle missing labels within training data. 

There are several interesting avenues of future work. For instance, while the straightforward \ac{HMLC} approach we take enjoys the virtue of being easy to implement and tune, it is possible that more sophisticated approaches, \eg{} using hierarchical features or dedicated classifiers, may garner even further improvements. Prior work using classic, non deep-learning approaches, explored these options~\cite{mccallum_improving_1998, cesa-bianchi_incremental_nodate,dumais_hierarchical_2000, cai_exploiting_nodate, vens_decision_2008}, and their insights should be applied today. Another important topic of future work should be on incorporating uncertainty within \ac{HMLC}. This would allow a model, when appropriate, to predict high confidence for non-leaf label predictions but lower confidence for leaf label predictions, enhancing its usefulness in deployment scenarios. Future work should also consider applications outside of \acp{CXR} both within and without medical imaging, \eg{} genomics or proteomics. Finally, one issue for further investigation is to better understand the implications of the annotation noise described by \citep{guendel_2019}, both for training and for evaluation. Relevant to this work, assessing label noise at higher levels of hierarchy should be an important focus going forward. 

\section*{Acknowledgements}

We thank the \ac{NCI} for access to \ac{NCI}'s data collected by the \ac{PLCO} Cancer Screening Trial. The statements	contained herein are solely ours and do not represent or imply concurrence or endorsement by \ac{NCI}. We also thank Chaochao Yan for help on pre-processing the \ac{PLCO} images and labels. Finally, we thank anonymous reviewers for their constructive comments and criticisms.

\bibliographystyle{model2-names.bst}\biboptions{authoryear}
\bibliography{haomin19_MIA}

\end{document}


\title{Supplementary materials for Deep Hierarchical Multi-label Classification of Chest X-ray Images}
\date{}

\maketitle 
\begin{table}
\begin{center}
\caption{AUC values for each disease pattern in the completely labelled experiments. Average values correspond to fine-grained disease patterns, i.e., the first 14 rows}
\label{tab:google}
\begin{tabular}{ ccccccccc } 
\hline
Disease Pattern & BR-leaf & BR-all & HLUP & HLCP & HLUP-finetune \\
\hline
Nodule &0.841&0.834&0.845&0.847&0.857\\
Mass &0.853&0.843&0.852&0.861&0.867\\
Distortion of Pulmonary Architecture &0.828&0.785&0.903&0.933&0.913\\
Pleural Based Mass &0.849&0.882&0.887&0.897&0.885\\
Granuloma &0.885&0.883&0.891&0.894&0.902\\
Fluid in Pleural Space &0.938&0.950&0.960&0.932&0.948\\
Hilar &0.872&0.849&0.836&0.868&0.879\\
Major Atelectasis &0.907&0.913&0.843&0.837&0.882\\
Infiltration &0.873&0.871&0.889&0.893&0.890\\
Scarring &0.843&0.841&0.842&0.840&0.852\\
Pleural Fibrosis &0.862&0.862&0.861&0.862&0.877\\
Bone Lesion &0.835&0.825&0.830&0.835&0.851\\
Cardiac Abnormality &0.928&0.923&0.912&0.925&0.928\\
COPD &0.879&0.878&0.861&0.876&0.883\\
Pulmonary Nodules and Masses &N/A&0.861&0.871&0.873&0.881\\
Opacity &N/A&0.872&0.883&0.886&0.885\\
Pleural Abnormality &N/A&0.866&0.862&0.864&0.878\\
Pulmonary Diseases &N/A&0.837&0.832&0.834&0.843\\
Abnormal &N/A&0.825&0.832&0.832&0.841\\
\hline
Average &0.871&0.867&0.872&0.879&0.887\\
\hline
\end{tabular}
\end{center}
\end{table}

\begin{table}
\begin{center}
\caption{AUC values conditioned on ``Abnormality'' for each disease pattern in the completely labelled experiments. Average values correspond to fine-grained disease patterns, i.e., the first 14 rows.}
\label{tab:google}
\begin{tabular}{ ccccccccc } 
\hline
Disease Pattern & BR-leaf & BR-all & HLUP & HLCP & HLUP-finetune \\
\hline
Nodule &0.761&0.751&0.760&0.783&0.788\\
Mass &0.764&0.750&0.751&0.767&0.783\\
Distortion of Pulmonary Architecture &0.693&0.696&0.827&0.880&0.852\\
Pleural Based Mass &0.757&0.801&0.804&0.832&0.816\\
Granuloma &0.853&0.852&0.859&0.870&0.871\\
Fluid in Pleural Space &0.894&0.918&0.921&0.890&0.907\\
Hilar &0.819&0.797&0.764&0.803&0.830\\
Major Atelectasis &0.817&0.838&0.721&0.725&0.785\\
Infiltration &0.790&0.791&0.796&0.818&0.818\\
Scarring &0.778&0.778&0.751&0.789&0.793\\
Pleural Fibrosis &0.793&0.802&0.788&0.799&0.821\\
Bone Lesion &0.808&0.788&0.781&0.818&0.830\\
Cardiac Abnormality &0.908&0.907&0.871&0.905&0.906\\
COPD &0.849&0.840&0.788&0.836&0.847\\
Pulmonary Nodules and Masses &N/A&0.815&0.822&0.839&0.839\\
Opacity &N/A&0.792&0.788&0.805&0.809\\
Pleural Abnormality &N/A&0.809&0.788&0.799&0.823\\
Pulmonary Diseases &N/A&0.790&0.736&0.786&0.796\\
\hline
Average &0.806&0.808&0.799&0.822&0.832\\
\hline
\end{tabular}
\end{center}
\end{table}

\begin{table}
\begin{center}
\caption{BR-leaf AUC values for each disease pattern in the incompletely labelled experiments. Average values correspond to fine-grained disease patterns, i.e., the first 12 rows.}
\label{tab:google}
\begin{tabular}{ ccccccccc } 
\hline
Disease Pattern & $\beta=0$ & $\beta=0.1$ & $\beta=0.2$ & $\beta=0.3$ & $\beta=0.4$ & $\beta=0.5$ & $\beta=0.6$ & $\beta=0.7$ \\
\hline
Nodule &0.799&0.790&0.780&0.785&0.779&0.766&0.744&0.749\\
Mass &0.780 &0.767&0.748&0.774&0.773&0.750&0.763&0.779\\
Pleural Based Mass &0.804&0.813&0.801&0.773&0.816&0.769&0.749&0.699\\
Granuloma &0.840&0.832&0.835&0.824&0.814&0.771&0.777&0.766\\
Fluid in Pleural Space &0.911&0.892&0.871&0.835&0.838&0.850&0.862&0.859\\
Hilar &0.814&0.814&0.784&0.772&0.783&0.775&0.754&0.739\\
Infiltration &0.824&0.808&0.812&0.813&0.792&0.778&0.782&0.723\\
Scarring &0.794&0.793&0.782&0.770&0.770&0.726&0.763&0.758\\
Pleural Fibrosis &0.818&0.818&0.790&0.787&0.780&0.765&0.727&0.724\\
Bone Lesion &0.768&0.744&0.768&0.752&0.745&0.741&0.743&0.733\\
Cardiac Abnormality &0.903&0.908&0.892&0.897&0.900&0.897&0.903&0.894\\
COPD &0.834&0.837&0.830&0.823&0.829&0.825&0.825&0.818\\
\hline
Average &0.824&0.818&0.808&0.800&0.802&0.784&0.783&0.770\\
\hline
\end{tabular}
\end{center}
\end{table}

\begin{table}
\begin{center}
\caption{BR-all AUC values for each disease pattern in the incompletely labelled experiments. Average values correspond to fine-grained disease patterns, i.e., the first 12 rows.}
\label{tab:google}
\begin{tabular}{ ccccccccc } 
\hline
Disease Pattern & $\beta=0$ & $\beta=0.1$ & $\beta=0.2$ & $\beta=0.3$ & $\beta=0.4$ & $\beta=0.5$ & $\beta=0.6$ & $\beta=0.7$ \\
\hline
Nodule &0.801&0.798&0.798&0.801&0.807&0.804&0.803&0.806\\
Mass &0.782&0.780&0.745&0.767&0.764&0.766&0.800&0.814\\
Pleural Based Mass &0.825&0.841&0.812&0.853&0.847&0.843&0.819&0.839\\
Granuloma &0.836&0.835&0.834&0.826&0.841&0.811&0.841&0.838\\
Fluid in Pleural Space &0.889&0.892&0.886&0.838&0.876&0.890&0.882&0.891\\
Hilar &0.823&0.795&0.802&0.761&0.795&0.784&0.759&0.742\\
Infiltration &0.825&0.811&0.820&0.840&0.883&0.826&0.849&0.786\\
Scarring &0.789&0.788&0.786&0.779&0.791&0.789&0.788&0.785\\
Pleural Fibrosis &0.816&0.810&0.809&0.805&0.818&0.801&0.827&0.814\\
Bone Lesion &0.753&0.760&0.760&0.744&0.747&0.750&0.747&0.762\\
Cardiac Abnormality &0.895&0.904&0.897&0.890&0.897&0.897&0.903&0.900\\
COPD &0.823&0.830&0.832&0.827&0.821&0.817&0.842&0.817\\
Pulmonary Nodules and Masses &0.814&0.812&0.812&0.806&0.813&0.806&0.813&0.808\\
Opacity &0.829&0.819&0.823&0.814&0.826&0.823&0.838&0.805\\
Pleural Abnormality &0.823&0.820&0.822&0.806&0.808&0.817&0.825&0.824\\
Pulmonary Diseases &0.797&0.798&0.794&0.788&0.790&0.798&0.798&0.791\\
Abnormal &0.788&0.787&0.790&0.781&0.787&0.785&0.784&0.781\\
\hline
Average &0.821&0.820&0.815&0.811&0.820&0.815&0.822&0.816\\
\hline
\end{tabular}
\end{center}
\end{table}

\begin{table}
\begin{center}
\caption{HLUP-finetune AUC values for each disease pattern in the incompletely labelled experiments. Average values correspond to fine-grained disease patterns, i.e., the first 12 rows.}
\label{tab:google}
\begin{tabular}{ ccccccccc } 
\hline
Disease Pattern & $\beta=0$ & $\beta=0.1$ & $\beta=0.2$ & $\beta=0.3$ & $\beta=0.4$ & $\beta=0.5$ & $\beta=0.6$ & $\beta=0.7$ \\
\hline
Nodule &0.815&0.814&0.812&0.821&0.821&0.817&0.821&0.831\\
Mass &0.791&0.797&0.792&0.783&0.753&0.792&0.779&0.797\\
Pleural Based Mass &0.827&0.853&0.822&0.833&0.863&0.823&0.802&0.858\\
Granuloma &0.854&0.853&0.856&0.855&0.858&0.846&0.849&0.858\\
Fluid in Pleural Space &0.919&0.916&0.893&0.905&0.916&0.940&0.881&0.939\\
Hilar &0.817&0.828&0.823&0.794&0.805&0.821&0.789&0.746\\
Infiltration &0.849&0.847&0.855&0.852&0.865&0.863&0.865&0.818\\
Scarring &0.812&0.808&0.811&0.812&0.815&0.808&0.802&0.814\\
Pleural Fibrosis &0.847&0.838&0.831&0.846&0.853&0.847&0.843&0.852\\
Bone Lesion &0.787&0.787&0.786&0.791&0.786&0.775&0.781&0.781\\
Cardiac Abnormality &0.911&0.909&0.911&0.908&0.913&0.907&0.897&0.911\\
COPD &0.843&0.838&0.836&0.845&0.844&0.833&0.823&0.827\\
Pulmonary Nodules and Masses &0.827&0.827&0.823&0.829&0.825&0.824&0.821&0.826\\
Opacity &0.851&0.840&0.852&0.842&0.855&0.851&0.842&0.845\\
Pleural Abnormality &0.852&0.841&0.844&0.847&0.847&0.850&0.835&0.845\\
Pulmonary Diseases &0.814&0.810&0.815&0.814&0.815&0.815&0.805&0.813\\
Abnormal &0.801&0.800&0.799&0.801&0.798&0.797&0.794&0.800\\
\hline
Average &0.839&0.841&0.836&0.837&0.841&0.839&0.828&0.836\\
\hline
\end{tabular}
\end{center}
\end{table}

\begin{center}
\begin{longtable}{|l|l|}
\caption{Included and excluded labels in PadChest}\\

\hline \multicolumn{1}{|c|}{\textbf{Included labels}} & \multicolumn{1}{c|}{\textbf{Excluded labels}} \\ \hline 
\endfirsthead

\multicolumn{2}{c}%
{{\bfseries \tablename\ \thetable{} -- continued from previous page}} \\
\hline \multicolumn{1}{|c|}{\textbf{Included labels}} & \multicolumn{1}{c|}{\textbf{Excluded labels}} \\ \hline 
\endhead

\hline \multicolumn{2}{|r|}{{Continued on next page}} \\ \hline
\endfoot

\hline \hline
\endlastfoot
sclerotic bone lesion (&central venous catheter\\
combined with ``blastic bone lesion'' and ``lytic bone lesion'', &empyema\\
and renamed as ``bone lesion'')&mediastinal mass\\
alveolar pattern&ascendent aortic elongation\\
consolidation&pectum excavatum\\
interstitial pattern&dextrocardia\\
ground glass pattern&segmental atelectasis\\
atelectasis&surgery humeral\\
lobar atelectasis&pleural effusion\\
laminar atelectasis&scoliosis\\
pleural thickening&air trapping\\
calcified pleural thickening&bronchovascular markings\\
apical pleural thickening&surgery neck\\
mass&hydropneumothorax\\
pulmonary mass&double J stent\\
granuloma (combined with ``calcified granuloma'')&total atelectasis\\
hilar enlargement&lipomatosis\\
vascular hilar enlargement&fibrotic band\\
adenopathy&rib fracture\\
nodule (combined with ``multiple nodules'')&soft tissue mass\\
aortic atheromatosis&dual chamber device\\
aortic elongation&azygoesophageal recess shift\\
descendent aortic elongation&pulmonary venous hypertension\\
aortic button enlargement&abscess\\
super aortic elongation&osteosynthesis material\\
infiltrates&osteopenia\\
cardiomegaly&external foreign body\\
bullas (renamed as ``COPD'')&mediastinal enlargement\\
&heart valve calcified\\
&central venous catheter via jugular vein\\
&dai\\
&fissure thickening\\
&lepidic adenocarcinoma\\
&single chamber device\\
&vertebral compression\\
&aortic endoprosthesis\\
&pulmonary hypertension\\
&heart insufficiency\\
&osteoporosis\\
&tracheostomy tube\\
&diaphragmatic eventration\\
&chest drain tube\\
&artificial heart valve\\
&air bronchogram\\
&artificial aortic heart valve\\
&catheter\\
&loculated fissural effusion\\
&superior mediastinal enlargement\\
&metal\\
&vascular redistribution\\
&thoracic cage deformation\\
&tuberculosis\\
&kyphosis\\
&hypoexpansion basal\\
&suboptimal study\\
&mastectomy\\
&esophagic dilatation\\
&fracture\\
&bone metastasis\\
&central venous catheter via subclavian vein\\
&lung vascular paucity\\
&chronic changes\\
&cavitation\\
&non axial articular degenerative changes\\
&calcified mediastinal adenopathy\\
&cervical rib\\
&hemidiaphragm elevation\\
&mediastinal shift\\
&hilar congestion\\
&calcified densities\\
&hiatal hernia\\
&pectum carinatum\\
&nephrostomy tube\\
&calcified fibroadenoma\\
&pneumomediastinum\\
&costophrenic angle blunting\\
&hypoexpansion\\
&calcified pleural plaques\\
&exclude\\
&humeral fracture\\
&pleural plaques\\
&pneumoperitoneo\\
&subacromial space narrowing\\
&flattened diaphragm\\
&pulmonary edema\\
&reticular interstitial pattern\\
&electrical device\\
&subcutaneous emphysema\\
&volume loss\\
&right sided aortic arch\\
&asbestosis signs\\
&surgery heart\\
&endoprosthesis\\
&humeral prosthesis\\
&loculated pleural effusion\\
&clavicle fracture\\
&abnormal foreign body\\
&round atelectasis\\
&artificial mitral heart valve\\
&gynecomastia\\
&ventriculoperitoneal drain tube\\
&NSG tube\\
&tracheal shift\\
&gastrostomy tube\\
&nipple shadow\\
&breast mass\\
&minor fissure thickening\\
&tuberculosis sequelae\\
&major fissure thickening\\
&mediastinic lipomatosis\\
&atypical pneumonia\\
&obesity\\
&pseudonodule\\
&air fluid level\\
&reticulonodular interstitial pattern\\
&pneumothorax\\
&bone cement\\
&endotracheal tube\\
&axial hyperostosis\\
&suture material\\
&respiratory distress\\
&Chilaiditi sign\\
&central venous catheter via umbilical vein\\
&pneumonia\\
&cyst\\
&post radiotherapy changes\\
&increased density\\
&pulmonary artery enlargement\\
&pleural mass\\
&pericardial effusion\\
&pulmonary artery hypertension\\
&calcified adenopathy\\
&reservoir central venous catheter\\
&miliary opacities\\
&kerley lines\\
&hyperinflated lung\\
&central vascular redistribution\\
&aortic aneurysm\\
&goiter\\
&mammary prosthesis\\
&prosthesis\\
&vertebral degenerative changes\\
&emphysema\\
&surgery\\
&pulmonary fibrosis\\
&pacemaker\\
&costochondral junction hypertrophy\\
&atelectasis basal\\
&callus rib fracture\\
&vertebral fracture\\
&lymphangitis carcinomatosa\\
&sternoclavicular junction hypertrophy\\
&COPD signs\\
&end on vessel\\
&surgery lung\\
&bronchiectasis\\
&lung metastasis\\
&vertebral anterior compression\\
&surgery breast\\
&azygos lobe\\
&sternotomy\\
\end{longtable}
\end{center}